\documentclass[final]{cvpr}

\usepackage{times}
\usepackage{epsfig}
\usepackage{graphicx}
\usepackage{amsmath}
\usepackage{bm}
\usepackage{amssymb}
\usepackage{float}

\usepackage{amsfonts}
\usepackage{xcolor}
\usepackage{gensymb}

\usepackage{url}            
\usepackage{booktabs}       
\usepackage{amsfonts}       
\usepackage{nicefrac}       
\usepackage{microtype}      

\usepackage{times}
\usepackage{epsfig}
\usepackage{graphicx}
\usepackage{amsmath}
\usepackage{amssymb}

\usepackage{units}
\usepackage{multirow}
\usepackage{bm}
\usepackage{bbm}
\usepackage{pifont}
\usepackage{xspace} 

\usepackage{graphicx,calc} 
\usepackage{color}    
\usepackage{ifthen}
\usepackage{paralist}
\usepackage{times}
\usepackage{longtable}
\usepackage{colortbl}
\usepackage{nomencl}
\usepackage{bm}
\usepackage{amssymb}
\usepackage{pifont}
\newcommand{\cmark}{\ding{51}}%
\newcommand{\xmark}{\ding{55}}%

\usepackage[pagebackref=true,breaklinks=true,colorlinks,bookmarks=false]{hyperref}

\def\cvprPaperID{5094} 
\def\confYear{CVPR 2021}

\newcommand{\todo}[1]{}
\renewcommand{\todo}[1]{{\color{red} TODO: {#1}}}

\begin{document}

\title{RSN: Range Sparse Net for Efficient, Accurate LiDAR 3D Object Detection}

\author{
  Pei Sun$^1$ \qquad Weiyue Wang$^1$ \qquad Yuning Chai$^1$ \qquad Gamaleldin Elsayed$^2$ \\
  \quad Alex Bewley$^2$ \qquad Xiao Zhang$^1$ \quad Cristian Sminchisescu$^2$ \quad Dragomir Anguelov$^1$ \\
  \\
  $^1$Waymo LLC, $^2$Google \\
  \texttt{peis@waymo.com} \\
}

\maketitle
\pagestyle{empty}
\thispagestyle{empty}

\begin{abstract}
The detection of 3D objects from LiDAR data is a critical component in most autonomous driving systems. Safe, high speed driving needs larger detection ranges, which are enabled by new LiDARs. These larger detection ranges require more efficient and accurate detection models.
Towards this goal, we propose Range Sparse Net (RSN) -- a simple, efficient, and accurate 3D object detector -- in order to tackle real time 3D object detection in this extended detection regime. 
RSN predicts foreground points from range images and applies sparse convolutions on the selected foreground points to detect objects.
The lightweight 2D convolutions on dense range images results in significantly fewer selected foreground points, thus enabling the later sparse convolutions in RSN to efficiently operate. Combining features from the range image further enhance detection accuracy.
RSN runs at more than 60 frames per second on a $150m\times150m$ detection region on Waymo Open Dataset (WOD) while being more accurate than previously published detectors. 
As of 11/2020, RSN is ranked first in the WOD leaderboard based on the APH/LEVEL\_1 metrics for LiDAR-based pedestrian and vehicle detection, while being several times faster than alternatives. 
\end{abstract}

\section{Introduction}

Concurrent with steady progress towards improving the accuracy and efficiency of 3D object detector algorithms \cite{yan2018second, qi2018frustum, meyer2019lasernet, lang2019pointpillars, zhou2019end, shi2019pointrcnn, ngiam2019starnet, shi2020pv, cheng2020improving, wang2020pillar, bewley2020range}, LiDAR sensor hardware has been improving in maximum range and fidelity, in order to meet the needs of safe, high speed driving. Some of the latest commercial LiDARs can sense up to 250m~\cite{packnet} and 300m~\cite{WaymoV5} in all directions around the vehicle. This large volume coverage places strong demands for efficient and accurate 3D detection methods.

\begin{figure}[th!]
    \centering
    \includegraphics[width=0.9\columnwidth]{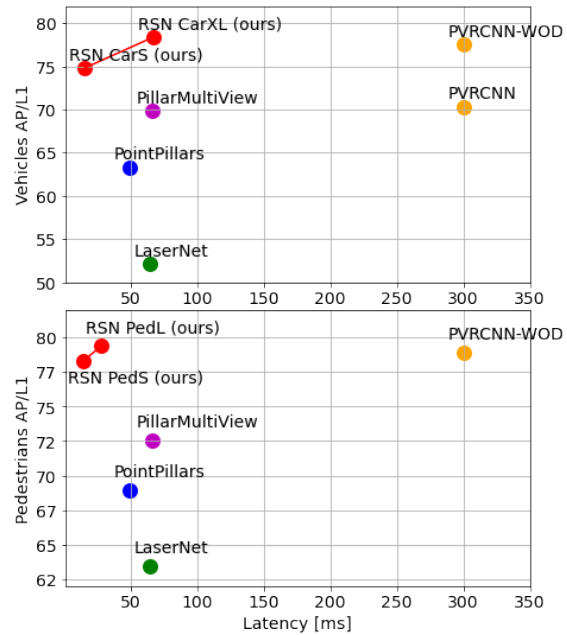}
    \caption{Accuracy (3D AP/L1 on WOD validation set) vs Latency (ms). RSN models significantly outperform others. See Table \ref{vehicle_result} and Table \ref{ped_result} for more details. \vspace{-0.5cm}}
    \label{fig:latency_ap}
\end{figure}

\textbf{Grid} based methods \cite{zhou2018voxelnet, lang2019pointpillars, zhou2019end, wang2020pillar, ge2020afdet} divide the 3D space into voxels or pillars, each of these being optionally encoded using PointNet \cite{qi2017pointnet}. Dense convolutions are applied on the grid to extract features. This approach is inefficient for large grids which are needed for long range sensing or small object detection. Sparse convolutions \cite{shi2020pv} scale better to large detection ranges but are usually slow due to the inefficiencies of applying to all points.
\textbf{Range images} are native, dense representations, suitable for processing point clouds captured by a single LiDAR. Range image based methods \cite{meyer2019lasernet, bewley2020range} perform convolutions directly over the range in order to extract point cloud features. 
Such models scale well with distance, but tend to perform less well in occlusion handling, accurate object localization, and for size estimation. A second stage, refining a set of initial candidate detections, can help mitigate some of these quality issues, at the expense of significant computational cost. 

To address the shortcomings of existing approaches, we introduce a novel 3D object detection model -- Range Sparse Net (RSN) -- which boosts the 3D detection accuracy and efficiency by combining the advantages of methods based on both dense range images and grids. RSN first applies a lightweight 2D convolutional network to efficiently learn semantic features from the high-resolution range image.  Unlike existing range image methods, which regress boxes directly from their underlying features, RSN is trained for high recall foreground segmentation. In a subsequent stage, sparse convolutions are applied only on the predicted foreground voxels and their learned range image features, in order to accurately regress 3D boxes. A configurable sparse convolution backbone and a customized CenterNet \cite{zhou2019objects} head designed for processing sparse voxels is introduced in order to enable end-to-end, efficient, accurate object detection without non-maximum-suppression. Figure \ref{fig:latency_ap} summarizes the main gains obtained with RSN models compared to others on the WOD validation set to demonstrate RSN's efficiency and accuracy.

RSN is a novel multi-view fusion method, as it transfers information from perspective view (range image) to the 3D view (sparse convolution on the foreground points). Its fusion approach differs from existing multi-view detection methods \cite{zhou2019end, wang2020pillar} in that 1) RSN's first stage directly operates on the high resolution range image while past approaches \cite{zhou2019end, wang2020pillar} perform voxelization (in a cylindrical or spherical coordinate system) that may lose some resolution, especially for small objects at a distance. 2) RSN's second stage processes only 3D points selected as foreground by the first stage, which yields improvements in both feature quality and efficiency.

RSN's design combines several insights that make the model very efficient. The initial stage is optimized to rapidly discriminate foreground from background points, a task that is simpler than full 3D object detection and allows a lightweight 2D image backbone to be applied to the range image at full resolution. The downstream sparse convolution processing is only applied on points that are likely to belong to a foreground object, which leads to additional, significant savings in compute. Furthermore, expensive postprocessing such as non-maximum suppression are eliminated by gathering local maxima center-ness points on the output, similar to CenterNet~\cite{zhou2019objects}. 

In this work, we make the following main contributions:  
\begin{compactitem}
    \item We propose a simple, efficient and accurate 3D LiDAR detection model RSN, which utilizes LiDAR range images to perform foreground object segmentation, followed by sparse convolutions to efficiently process the segmented foreground points to detect objects.
    \item We propose a simple yet effective temporal fusion strategy in RSN with little additional inference cost.
    \item In experiments on the Waymo Open Dataset \cite{sun2020scalability} (WOD), we demonstrate the state of art accuracy and efficiency for vehicle and pedestrian detection. Experiments on an internal dataset further demonstrate RSN's scalability for long-range object detection.  
    \item We conduct ablation studies to examine the effectiveness of range image features and the impact of aspects like foreground point selection thresholds, or end-to-end model training, on both latency and accuracy.
\end{compactitem}


\section{Related Work}

\subsection{LiDAR Data Representation} 
The are four major LiDAR data representations for 3D object detection including voxel grids, point sets, range images, and hybrids. 

\noindent\textbf{Voxel grid based methods.} 3D points are divided into a grid of voxels. Each voxel is encoded with hand-crafted  metrics such as voxel feature means and covariances. 
Vote3Deep~\cite{engelcke2017vote3deep} was the first to apply a deep network composed of sparse 3D convolutions to 3D detection. They also proposed an $L_1$ penalty to favour sparsity in deeper layers.
The voxels can be scattered to a pseudo-image which can be processed by standard image detection architectures. MV3D \cite{chen2017mv3d}, PIXOR \cite{yang2018pixor} and Complex YOLO \cite{simon2018complex} are  notable models based on this approach. 
VoxelNet \cite{zhou2018voxelnet} applied PointNet \cite{qi2017pointnet} in each voxel to avoid handcrafted voxel features. PointPillars \cite{lang2019pointpillars} introduced 2D pillar to replace 3D voxel to boost model efficiency. For small enough 3D voxel sizes, the PointNet can be removed if 3D sparse convolutions are used. Notable examples based on this approach include Second \cite{yan2018second} and PVRCNN \cite{shi2020pv}.


There are three major drawbacks to voxel based methods. 1) Voxel size is constant at all ranges which limits the model's capability at distance and usually needs larger receptive fields. 2) The requirement of a full 3D grid poses a limitation for long-range, since both complexity and memory consumption scale quadratically or cubically with the range. Sparse convolutions can be applied to improve scalability but is usually still limited by the large number of voxels. 3) The voxel representation has a limited resolution due to the scalability issue mentioned above.

\noindent\textbf{Point set based methods.} This line of methods treats point clouds as unordered sets. Most approaches are based on the seminal  PointNet and variants \cite{qi2017pointnet, qi2017pointnet++}. FPointNet\cite{qi2018frustum} detects objects from a cropped point cloud given by 2D proposals obtained from images; PointRCNN\cite{shi2019pointrcnn} proposes objects directly from each point; STD \cite{yang2019std} relies on a sparse to dense strategy for better proposal refinement; DeepHough \cite{qi2019deepvote} explores deep hough voting to better group points before generating box proposals. Although these methods have the potential to scale better with range, they lag behind the quality of voxel methods. Moreover, they require nearest neighbor search for the input, scaling with the number of points, which can be costly.

\noindent\textbf{Range image based methods.} 
Despite being a native and dense representation for 3D points captured from a single view-point e.g. LiDAR, prior work on using 2D range images is not extensive. LaserNet \cite{meyer2019lasernet} applied a traditional 2D convolution network to range image to regress boxes directly. RCD-RCNN \cite{bewley2020range} pursued range conditioned dilation to augment traditional 2D convolutions, followed by a second stage to refine the proposed range-image boxes which is also used by Range-RCNN \cite{liang2021rangercnn}. 
Features learned from range images alone are very efficient when performing 2D convolutions on 2D images but aren't that good at handling occlusions, for accurate object localization, and for size regression, which usually requires more expressive 3D features.

\noindent\textbf{Hybrid methods.} MultiView \cite{zhou2019end} fuses features learned from voxels in both spherical and Cartesian coordinates to mitigate the limited long-range receptive fields resulting from the fixed-voxel discretization in grid based methods. Pillar-MultiView \cite{wang2020pillar} improves \cite{zhou2019end} by further projecting fused spherical and cartesian features to bird-eye views followed by additional convolution processing to produce stronger features. These methods face similar scalability issues as voxel approaches.

\subsection{Object Detection Architectures} Typical two-stage detectors \cite{girshick2014rcnn, girshick2015fast, ren2016faster, dai2016rfcn} generate a sparse set of regions of interest (RoIs) and classify each of them by a network. PointRCNN \cite{shi2019pointrcnn}, PVRCNN \cite{shi2020pv}, RCD-RCNN \cite{bewley2020range} share similar architectures with Faster-RCNN but rely on different region proposal networks designed for different point cloud representations. Single-stage detectors were popularized by the introduction of YOLO \cite{redmon2016you}, SSD \cite{liu2016ssd} and RetinaNet \cite{lin2017focal}. Similar architectures are used to design single stage 3D point cloud methods \cite{zhou2018voxelnet, lang2019pointpillars, yan2018second, zhou2019end, wang2020pillar}. These achieve competitive accuracy compared to two stage methods such as PVRCNN \cite{shi2020pv} but have much lower latency. Keypoint-based architectures such as CornerNet \cite{law2018cornernet} and CenterNet \cite{zhou2019objects} enable end to end training without non-maximum-suppression. AFDet \cite{ge2020afdet} applies a CenterNet-style detection head to a PointPillars-like detector for 3D point clouds.
Our proposed RSN method also relies on two stages. However the first stage performs segmentation rather than box proposal estimation, and the second stage detects objects from segmented foreground points rather than performing RoI refinement. RSN adapts the CenterNet detection head to sparse voxels.

\section{Range Sparse Net}
\label{sec:rsn}
\begin{figure}[t!]
    \centering
    \includegraphics[width=0.98\columnwidth]{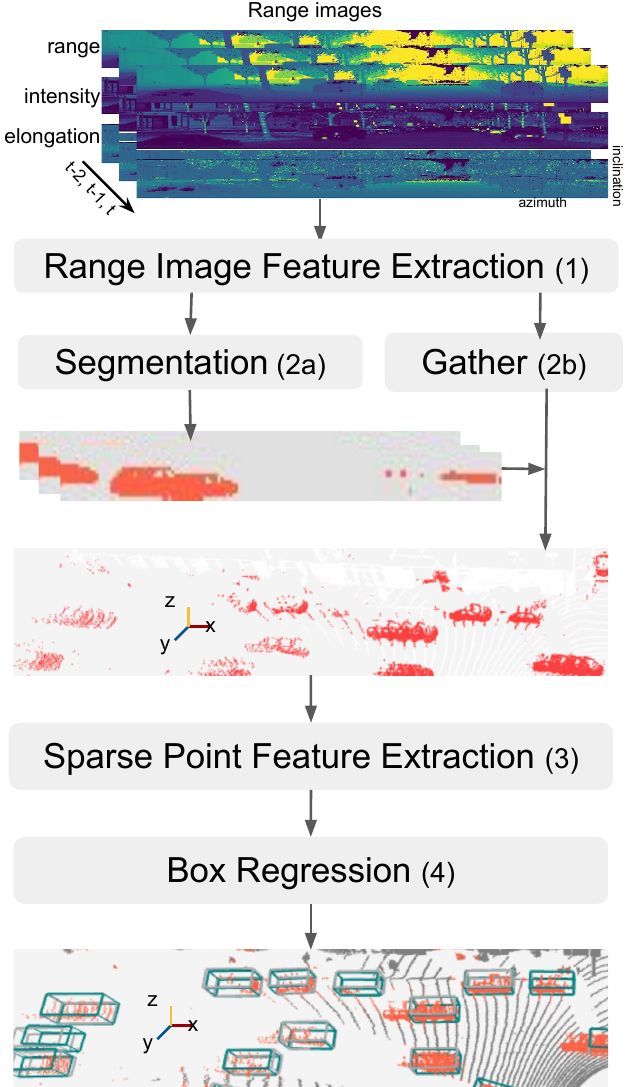}
    \hfill
    \caption{(Best viewed in color) Range Sparse Net object detection architecture. The net consists of five components: 1) Range image feature extraction: a 2D convolution net on range images to extract associated image features. 2) Foreground point selection: foreground points are segmented on range images in 2a); together with the learned range image features, they are gathered to sparse points in 2b). 3) Sparse point feature extraction: per-point features are extracted on the selected foreground points by applying sparse convolutions. 4) A sparse CenterNet head to regress boxes. Red points are selected foreground points. Light gray boxes are ground truths. Teal boxes are detection results.\vspace{-0.6cm}}
    \label{fig:arch}
\end{figure}

The main contribution of this work is the Range Sparse Net (RSN) architecture (Fig. \ref{fig:arch}). RSN accepts raw LiDAR range images \cite{sun2020scalability} as input to an efficient 2D convolution backbone that extracts range image features. A segmentation head is added to process range image features. This segments background and foreground points, with the foreground being points inside ground truth objects. Unlike traditional semantic segmentation, recall is emphasized over high precision in this network. We select foreground points based on the segmentation result.
The selected foreground points are further voxelized and fed into a sparse convolution network. These sparse convolutions are very efficient because we only need to operate on a small number of foreground points. At the end, we apply a modified CenterNet \cite{zhou2019objects} head to regress 3D boxes efficiently without non-maximum-suppression.

\subsection{Range Image Feature Extraction (RIFE)}
\label{sec:ri_fe}

Range images are a native dense representation of the data captured by LiDAR sensors. Our input range images contain range, intensity and elongation channels, where range is the distance from LiDAR to the point at the time the point is collected, while intensity and elongation are LiDAR return properties which can be replaced or augmented with other LiDAR specific signals. The channel values of the input range images are normalized by clipping and rescaling to $[0, 1]$. 

\begin{figure}[th!]
    \centering
    \includegraphics[width=0.8\columnwidth]{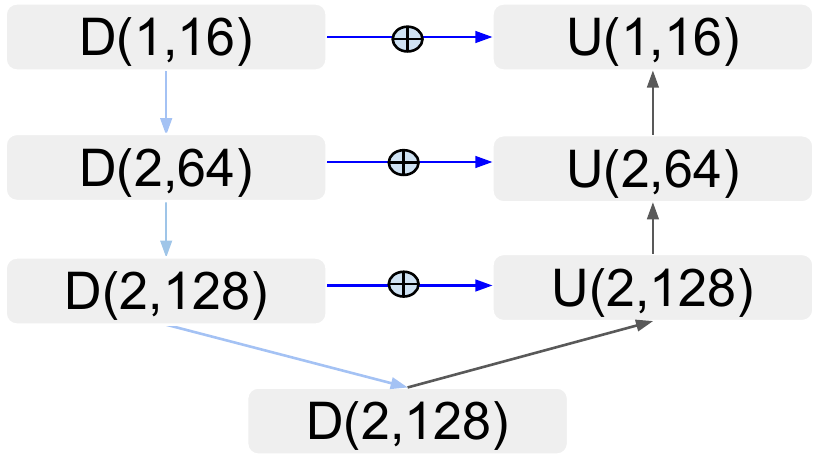}
    \caption{Range image U-Net feature extractor to compute high level semantic range features. See section \ref{sec:ri_fe} for details. \vspace{-0.5cm}}
    \label{fig:unet}
\end{figure}

A 2D convolution net is applied on the range image to simultaneously learn range image features and for foreground segmentation. 
We adopt a lightweight U-Net \cite{ronneberger2015u} with its structure shown in Fig. \ref{fig:unet}. Each $D(L,C)$ downsampling block contains $L$ resnet \cite{HeResnet} blocks each with $C$ output channels. Within each block the first has stride 2. Each $U(L, C)$ block contains 1 upsampling layer and $L$ resnet blocks. All resnet blocks have stride 1. The upsampling layer consists of a $1\times1$ convolution followed by a bilinear interpolation.

\subsection{Foreground Point Selection}
\label{sec:seg}
To maximize efficiency through sparsity in the downstream processing, the output of this 2D convolutional network is an ideal place to reduce the input data cloud to only points most likely to belong to an object. Here, a $1\times1$ convolutional layer performs pixelwise foreground classification on the learned range image features from \ref{sec:ri_fe}. This layer is trained using the focal loss \cite{lin2017focal} with ground truth labels derived from 3d bounding boxes by checking whether the corresponding pixel point is in any box.
\begin{align}
\label{eq:seg_loss}
    L_{\textrm{seg}} = \frac{1}{P}\sum_{i}{L_{i}},
\end{align}
$P$ is the total number of valid range image pixels. $L_i$ is the focal loss for point $i$. Points with foreground score $s_i$ greater than a threshold $\gamma$ are selected.
As false positives can be removed in the later sparse point feature extraction phase (\S\ref{sec:spfe}) but false negatives cannot be recovered, the foreground threshold is selected to achieve high recall and acceptable precision.

\subsection{Sparse Point Feature Extraction (SPFE)}
\label{sec:spfe}
We apply dynamic voxelization \cite{zhou2019end} on the selected foreground points. Similar to PointPillars \cite{lang2019pointpillars}, we append each point $\boldsymbol{p}$ with $\boldsymbol{p - m}, \textbf{\textrm{var}}, \boldsymbol{p - c}$ where $\boldsymbol{m}$, $\textbf{\textrm{var}}$ is the arithmetic mean and covariance of each voxel, $\boldsymbol{c}$ is the voxel center. Voxel sizes are denoted as $\Delta_{x,y,z}$ along each dimension. When using a pillar style voxelization where 2D sparse convolution is applied, $\Delta_z$ is set to $+\infty$. The selected foreground points are encoded into sparse voxel features which can optionally be further processed by a PointNet \cite{qi2017pointnet}.

\begin{figure}[th!]
    \centering
    \includegraphics[width=1\columnwidth]{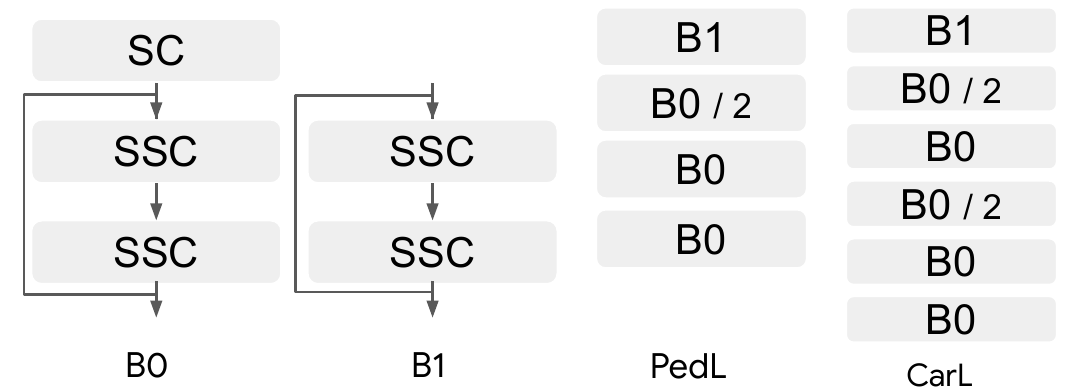}
    \caption{SPFE building blocks and example net architectures. See section \ref{sec:spfe} for usage details. SC denotes 3x3 or 3x3x3 sparse convolution \cite{graham2017submanifold} with stride 1 or 2. SSC denotes 3x3 or 3x3x3 submanifold sparse convolution. PedL (2D) and CarL (2D) are the large pedestrian and vehicle SPFEs. $/2$ denotes stride 2. \vspace{-0.5cm}}
    \label{fig:sconv}
\end{figure}

A 2D or 3D sparse convolution network (for pillar style, or 3D type voxelization, respectively) is applied on the sparse voxels. Fig. \ref{fig:sconv} illustrates the net building blocks and example net architectures used for vehicle and pedestrian detection. More network architecture details can be found in the Appendix \ref{app:spfe_detail}.


\subsection{Box Regression}
\label{sec:br}
\begin{figure}[th!]
    \centering
    \includegraphics[width=0.8\columnwidth]{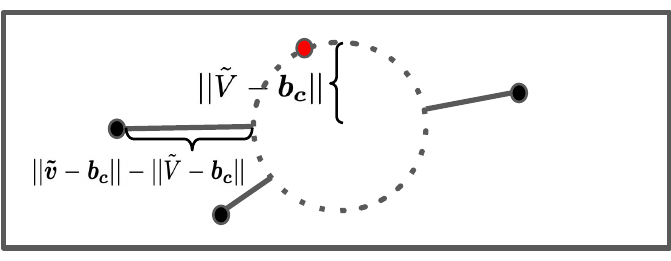}
    \caption{RSN centerness heatmap computation. The heatmap value is computed by the distance between the point and the circle placed at the box center, with radius being the distance from the box center to the closest point (red point). \vspace{-0.4cm}}
    \label{fig:center_heat}
\end{figure}

We use a modified CenterNet \cite{zhou2019objects, ge2020afdet} head to regress boxes from point features efficiently. The feature map consists of voxelized coordinates $V = \{\boldsymbol{v} | \boldsymbol{v} \in \mathbb{N}_{0}^d\}$, where $d\in\{2,3\}$ depending on whether 2D or 3D SPFE is used. We scale and shift it back to the raw point Cartesian coordinate as $\Tilde{V} = \{\boldsymbol{\Tilde{v}} | \boldsymbol{\Tilde{v}} \in R^d\}$. The ground truth heatmap for any $\boldsymbol{\Tilde{v}} \in \Tilde{V}$ is computed as $h = \max\{\exp(-\frac{||\boldsymbol{\Tilde{v}} - \boldsymbol{b_c}|| - ||\Tilde{V} - \boldsymbol{b_c}||}{\sigma^2}) | \boldsymbol{b_c} \in B_c(\boldsymbol{\Tilde{v}})\}$ where $B_c(\boldsymbol{\Tilde{v}})$ is the set of centers of the boxes that contain $\boldsymbol{\Tilde{v}}$. $h = 0$ if $|B_c(\boldsymbol{\Tilde{v}})| = 0$. This is illustrated in Fig. \ref{fig:center_heat}. $\sigma$ is a per class constant. We use a single fully connected layer to predict heatmap and box parameters. The heatmap is regressed with a penalty-reduced focal loss \cite{zhou2019objects, lin2017focal}.
\begin{align}
\label{eq:hm_loss}
\begin{split}
    L_{\textrm{hm}} = -\frac{1}{N}\sum_{\Tilde{p}}\{
    (1 - \Tilde{h})^\alpha\log(\Tilde{h})I_{h > 1 - \epsilon} + \\ (1-h)^\beta\Tilde{h}^\alpha\log(1-\Tilde{h})I_{h \leq 1 - \epsilon}\},
\end{split}
\end{align}
where $\Tilde{h}$ and $h$ are the predicted and ground truth heatmap values respectively. $\epsilon$, added for numerical stability, is set to $1e-3$. We use $\alpha=2$ and $\beta=4$ in all experiments, following \cite{zhou2019objects, law2018cornernet}.

The 3D boxes are parameterized as $\boldsymbol{b} = \{d_x, d_y, d_z, l, w, h, \theta\}$ where $d_x, d_y, d_z$ are the box center offsets relative to the voxel centers. Note that $d_z$ is the same as the absolute box $z$ center if 2D point feature extraction backbone is used (see \ref{sec:spfe}). $l, w, h, \theta$ are box length, width, height and box heading. A bin loss \cite{shi2019pointrcnn} is applied to regress heading $\theta$. The other box parameters are directly regressed under smooth L1 losses. IoU loss \cite{zhou2019iou} is added to further boost box regression accuracy. Box regression losses are only active on the feature map pixels that have ground truth heatmap values greater than a threshold $\delta_1$.
\begin{align}
\label{eq:bin_loss}
L_{\theta_i} &= L_{bin}(\theta_i, \Tilde{\theta}_i), \\
L_{\boldsymbol{b_{i}} \backslash \theta_{i}} &= \textrm{SmoothL1}(\boldsymbol{b_{i}} \backslash \theta_{i} -  \boldsymbol{\Tilde{b_{i}}} \backslash \Tilde{\theta}_i), \\
L_{\textrm{box}} &= \frac{1}{N} \sum_i {(L_{\theta_i} + L_{\boldsymbol{b_i} \backslash \theta_{i}} + L_{\textrm{iou}_{i}}) I_{h_i > \delta_1}},
\end{align}
where $\Tilde{b}_i$, $b_i$ are the predicted and ground truth box parameters respectively, $\Tilde{\theta}_i$, $\theta_i$ are the predicted and ground truth box heading respectively. $h_i$ is the ground truth heatmap value computed at feature map pixel $i$.

The net is trained end to end with the total loss defined as
\begin{equation}
\label{eq:total_loss}
L  = \lambda_1 L_{\textrm{seg}} + \lambda_2 L_{\textrm{hm}} + L_{\textrm{box}}
\end{equation}
\vspace{-0.5cm}
\label{sec:box_reg}

We run a sparse submanifold max pooling operation on the sparse feature map voxels that have heatmap prediction greater than a threshold $\delta_2$. Boxes corresponding to local maximum heatmap predictions are selected.

\subsection{Temporal Fusion}
\label{sec:temporal}
Existing range image based detection methods \cite{meyer2019lasernet} \cite{bewley2020range} are not temporal fusion friendly as range images are constructed while the self-driving-car (SDC) moves. Stacking range images directly gives little benefit for detection performance due to ego-motion. Removing ego-motion from the range images is not optimal because range reconstructions at a different frame results in non-trivial quantization errors.  

Temporal RSN takes a sequence of temporal invariant range images as input as shown in Fig. \ref{fig:arch}. RIFE is applied on each range image to segment foreground points and extract range image features. Then we transform all the selected points to the latest frame to remove ego-motion. During the SPFE phase, we append to each point voxel features computed from its own frame instead of all frames. This works better because it avoids mixing points from different frames together during voxelization. In addition, we append the time difference in seconds w.r.t. the latest frame to each point to differentiate points from different frames. The selected foreground points from all frames are processed by the SPFE backbone same as single frame models.

\section{Experiments}
We introduce the RSN implementation details and illustrate its efficiency and accuracy in multiple experiments. Ablation studies are conducted to understand the importance of various RSN components. 

\subsection{Waymo Open Dataset}
We primarily benchmark on the challenging Waymo Open Dataset (WOD)~\cite{sun2020scalability}. WOD released its raw data in high quality range image format directly, which makes it a better fit for building range image models. 
The dataset contains 1150 sequences in total, split into 798 training, 202 validation, and 150 test. Each sequence contains about 200 frames, where each frame captures the full 360 degrees around the ego-vehicle that results in a range image of a dimension $64\times2650$ pixels. The dataset has one long range LiDAR with range capped at 75 meters and four near range LiDARs. We only used data from the long range LiDAR but still evaluated our results on full range. In practice, RSN can be adapted to accept multiple LiDAR images as inputs.

\subsection{Implementation Details}
RSN is implemented in the Tensorflow framework \cite{tensorflow} with sparse convolution implementation similar as \cite{yan2018second}. Pedestrians and vehicles are trained separately with different SPFEs (\S\ref{sec:spfe}). They share the same RIFE (\S\ref{sec:ri_fe}). We show results from 3 vehicle models CarS, CarL, CarXL and 2 pedestrian models PedS, PedL with network details described in \S\ref{sec:spfe} and Appendix \ref{app:spfe_detail}. Each model can be trained with single frame input (e.g. CarS\_1f) or 3 frame input (e.g. CarS\_3f). The input images are normalized by $\min(v, m) / m$ where $v$ is range, intensity and elongation, $m$ is 79.5, 2.0, 2.0 respectively. The last return is picked if there are multiple laser returns.

The foreground score cutoff $\gamma$ in \S\ref{sec:seg} is set to 0.15 for vehicle and 0.1 for pedestrian. The segmentation loss weight $\lambda_1$ in Eq.\ref{eq:total_loss} is set to 400. The voxelization region is $[-79.5m, 79.5m]\times[-79.5m, 79.5m]\times[-5m, 5m]$. The voxel sizes are set to 0.2 meter and 0.1 meter for vehicle model and pedestrian model respectively. Per object $\sigma$ in the heatmap computation is set to 1.0 for vehicle and 0.5 for pedestrian. The heatmap loss weight $\lambda_2$ is set to 4 in Eq. \ref{eq:total_loss}. The heatmap thresholds $\delta_1$, $\delta_2$ in \S\ref{sec:box_reg} are both set to 0.2. We use 12 and 4 bins in the heading bin loss in Eq. \ref{eq:bin_loss} for heading regression for vehicle and pedestrian, respectively. 


\begin{table*}[th!]
\small{
\begin{center}
\begin{tabular}{l|c|cc|cc|ccc}
\toprule
\multirow{2}{*}{Method} &
{Latency} &
\multicolumn{2}{c|}{AP/APH L1} &
\multicolumn{2}{c|}{AP/APH L2} &
\multicolumn{3}{c}{AP/APH L1 3D by distance} \\
& (ms) & BEV & 3D & BEV & 3D & $<$30m & 30-50m & $>50$m \\
\midrule
LaserNet CVPR'19 \cite{meyer2019lasernet} * & 64.3 & 71.2/67.7 & 52.1/50.1 & - & - & 70.9/68.7 & 52.9/51.4 &29.6/28.6 \\
P.Pillars CVPR'19\cite{lang2019pointpillars} \dag & 49.0 & 82.5/81.5 & 63.3/62.7 & 73.9/72.9 & 55.2/54.7 & 84.9/84.4 & 59.2/58.6 & 35.8/35.2 \\
PillarMultiView ECCV'20\cite{wang2020pillar} & 66.7\ddag & 87.1/- & 69.8/- & - & - & 88.5/- & 66.5/- & 42.9/- \\
PVRCNN CVPR'20\cite{shi2020pv} & - & 83.0/82.1 & 70.3/69.7 & 77.5/76.6 & 65.4/64.8 & 91.9/91.3 & 69.2/68.5 & 42.2/41.3 \\
PVRCNN WOD'20\cite{shi2020pvreport} & 300 \P &  - & 77.5/76.9 & - & 68.7/68.2 & - &- &-\\
RCD CORL'20 \cite{bewley2020range} & - & 82.1/83.4 & 69.0/68.5 & - & - & 87.2/86.8 & 66.5/66.1& 44.5/44.0 \\
\midrule
RSN CarS\_1f (Ours) & - & 86.7/86.0 & 70.5/70.0 & 77.5/76.8 & 63.0/62.6 & 90.8/90.4  & 67.8/67.3 & 45.4/44.9 \\
RSN CarS\_3f (Ours) & \textbf{15.5} & 88.1/87.4 & 74.8/74.4 & 80.8/80.2 & 65.8/65.4 & 92.0/91.6  & 73.0/72.5 & 51.8/51.2 \\
RSN CarL\_1f (Ours) & - & 88.5/87.9 & 75.1/74.6 & 81.2/80.6 & 66.0/65.5 & 91.8/91.4 & 73.5/73.1 & 53.1/52.5 \\
RSN CarL\_3f (Ours) & 25.4 & 91.0/90.3 & 75.7/75.4 & 82.1/81.6 & 68.6/68.1 & 92.1/91.7 & 74.6/74.1 & 56.1/55.4  \\
RSN CarXL\_3f (Ours) & 67.5 & \textbf{91.3/90.8} & \textbf{78.4/78.1} & \textbf{82.6/82.2} & \textbf{69.5/69.1} & 92.1/91.7 & 77.0/76.6 & 57.5/57.1\\
\bottomrule
\end{tabular}
\end{center}
}
\caption{Performance comparisons on the Waymo Open Dataset \textit{validation set} for vehicle detection. (*) is re-implemented by \cite{bewley2020range}. (\dag) is our re-implementation with flip and rotation data augmentation following PointPillar setting in \cite{sun2020scalability} which is better than other PointPillars re-implementations such as \cite{zhou2019end}. (\ddag) is from \cite{wang2020pillar}. \P is obtained privately from PVRCNN authors who benchmarked on Titan RTX. All the other latency numbers are obtained based on our own implementations on Tesla V100 GPUs. They are averaged on 10 scenes, each has more than 100 vehicles. }\vspace{-0.1cm}
\label{vehicle_result}
\end{table*}

\begin{table*}[th!]
\small{
\begin{center}
\begin{tabular}{l|c|cc|cc|ccc}
\toprule
\multirow{2}{*}{Method} &
{Latency} &
\multicolumn{2}{c|}{AP/APH L1} &
\multicolumn{2}{c|}{AP/APH L2} &
\multicolumn{3}{c}{AP/APH L1 3D by distance} \\
& (ms) & BEV & 3D & BEV & 3D & $<30$m & 30-50m & $>50$m \\
\midrule

LaserNet CVPR'19\cite{meyer2019lasernet}* & 64.3 & 70.0/- & 63.4/- & - & - & 73.5/- & 61.6/- & 42.7/- \\
P.Pillars CVPR'19\cite{lang2019pointpillars}\dag & 49.0 & 76.0/62.0 & 68.9/56.6 & 67.2/54.6 & 60.0/49.1 & 76.7/64.3 & 66.9/54.3 & 52.9/40.5 \\ 
PillarMultiView ECCV'20\cite{wang2020pillar} & 66.7\ddag & 78.5/-& 72.5/- & - & - & 79.3/-----&72.1/-----& 56.8/----- \\
PVRCNN WOD'20\cite{shi2020pvreport} & 300 \P &  - & 78.9/75.1 & - & 69.8/66.4 & - & - & -\\
\midrule

RSN PedS\_1f (Ours) & - & 80.7/74.9 & 74.8/69.6 & 71.2/65.9  & 65.4/60.7 & 81.4/77.4 & 72.8/66.8& 59.0/50.6 \\
RSN PedS\_3f (Ours) & \textbf{14.4} & 84.2/80.7 & 78.3/75.2 & 74.8/71.6 & 68.9/66.1 & 81.7/78.8 & 74.4/71.3 & 64.9/61.5 \\
RSN PedL\_1f (Ours) & - & 83.4/77.6 & 77.8/72.7 & 73.9/68.6 & 68.3/63.7 & 83.9/79.7 & 74.1/68.2 & 62.1/54.1 \\
RSN PedL\_3f (Ours) & 28.2 & \textbf{85.0/81.4} & \textbf{79.4/76.2} & \textbf{75.5/72.2} & \textbf{69.9/67.0} & 84.5/81.5 & 78.1/74.7 & 68.5/65.0 \\ 
\bottomrule
\end{tabular}
\end{center}
}
\caption{Performance comparison on the Waymo Open Dataset \textit{validation set} for pedestrian detection. See Table \ref{vehicle_result} for details on ways to obtaining latency numbers. All the latency of our models are averaged on 10 scenes, each has more than 50 pedestrians.}\vspace{-0.1cm}
\label{ped_result}
\end{table*}

\begin{table*}[th!]
\small{
\begin{center}
\begin{tabular}{l|cccc|cccc}
\toprule
\multirow{2}{*}{Method} &
\multicolumn{4}{c|}{AP/APH L1 3D} &
\multicolumn{4}{c}{AP/APH L2 3D} \\
& Overall & $<30$m & 30-50m & $>50$m & Overall & $<30$m & 30-50m & $>50$m \\
\midrule
& \multicolumn{8}{c}{VEHICLE} \\
P.Pillars CVPR'19\cite{lang2019pointpillars} \dag & 68.6/68.1 & 87.2/86.7  & 65.5/64.9 & 40.9/40.2 & 60.5/60.1 & 85.9/85.4 & 58.9/58.3 & 31.3/30.8 \\ 
PVRCNN Ensem WOD'20 \cite{shi2020pvreport} & 81.1/80.6 & 93.4/93.0 & 80.1/79.6 & 61.2/60.5 & \textbf{73.7/73.2} & 92.5/92.0 & 74.0/73.5 & 49.3/48.6 \\
RSN CarXL\_3f (Ours) & \textbf{80.7/80.3} & 92.2/91.9 & 79.1/78.7 & 63.0/62.5 & \textbf{71.9/71.6} & 91.5/91.1 & 71.4/71.1 & 49.9/49.5\\
RSN CarXL\_3f Ensem (Ours) & \textbf{81.4/81.0} & 92.4/92.0 & 80.2/79.8 & 64.7/64.1 & 72.8/72.4 & 91.5/91.1 & 74.2/73.8 & 51.3/50.8  \\
\toprule
\toprule
& \multicolumn{8}{c}{PEDESTRIAN} \\
P.Pillars CVPR'19\cite{lang2019pointpillars} \dag & 68.0/55.5 & 76.0/63.5  & 66.8/54.1 & 54.3/42.1 & 61.4/50.1 & 73.4/61.2 & 61.5/49.8 & 43.9/34.0 \\
PVRCNN Ensem WOD'20 \cite{shi2020pvreport} & 80.3/76.3 & 86.7/82.9 & 78.9/74.8 & 70.5/66.4 & 74.0/70.2 & 84.8/80.9 & 73.6/69.6 & 59.2/55.5 \\
RSN PedL\_3f (Ours) & \textbf{78.9/75.6} & 85.5/82.4 & 77.5/74.2 & 67.3/64.1 & \textbf{70.7/67.8} & 81.9/78.9 & 70.3/67.3 & 55.8/53.0 \\
RSN PedL\_3f Ensem (Ours) & \textbf{82.4/78.0} & 89.1/85.0 & 81.1/76.8  & 70.7/66.3  & \textbf{74.7/70.7} & 86.0/82.0 & 74.6/70.6 & 58.7/54.8 \\
\bottomrule
\end{tabular}
\end{center}
}
\caption{Performance comparison on the Waymo Open Dataset \textit{test set}. (\dag) is our re-implementation as described in Table \ref{vehicle_result}. 'Ensem' is short for ensemble. See Appendix \ref{app:ensem} for  details.}\vspace{-0.4cm}
\label{test_set_result}
\end{table*}

\begin{figure*}[t!]
    \centering
    \includegraphics[height=0.46\textwidth]{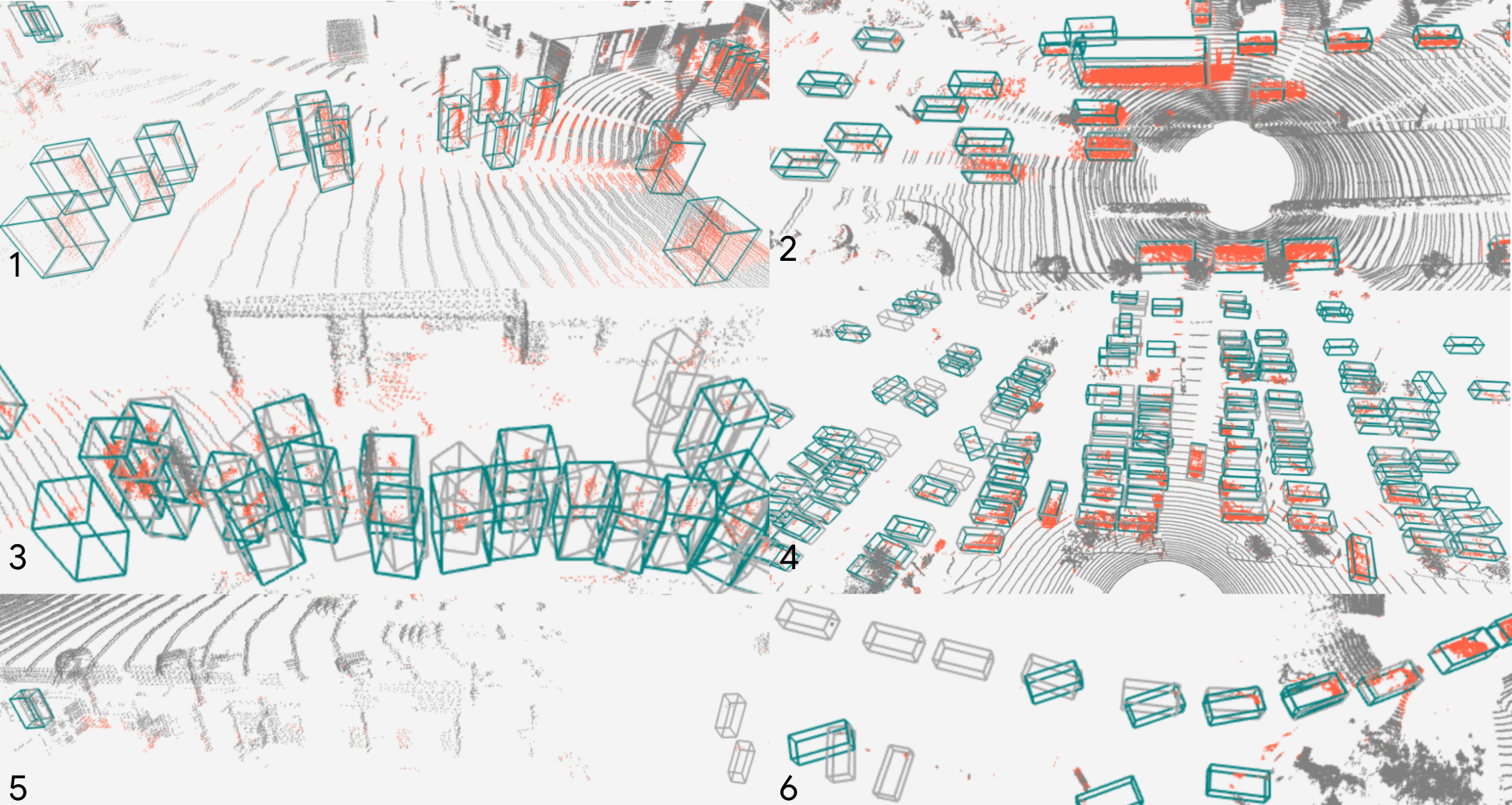}
    \hfill
    \caption{(Best viewed in color) Example pedestrian and vehicle detection results of CarS\_3f and PedS\_3f on the Waymo Open Dataset validation set. Light gray boxes are ground-truth and teal boxes are our prediction results. Red points are selected foreground points. Ex 1, 2: RSN performs well when objects are close and mostly visible. Both vehicles and pedestrians are predicted with high accuracy, including dynamic vehicles, large vehicles. Ex 3, 4: RSN handles large crowds with severe occlusion with few false positives and false negatives. Many of the false-negatives in Ex 4 have very few points in the ground-truth boxes. Ex 5, 6: Typical failures of RSN are for distant or heavily occluded objects and having very few points observed. \vspace{-0.3cm}}
    \label{fig:ex}
\end{figure*}

\subsection{Training and Inference}
\label{sec:train_and_inf}
RSN is trained from scratch end-to-end using an ADAM optimizer \cite{kingma2014adam} on Tesla V100 GPUs. Different SPFE backbones are trained with the maximum possible batch sizes to fit the net in GPU memory. Single frame models are trained on 8 GPUs. 3-frame temporal models are trained on 16 GPUs. We adopted the cosine learning rate decay, with initial learning rate set to 0.006, 5k warmup steps starting at 0.003, 120k steps in total. We observed that accuracy metrics such as AP fluctuate during training because the points selected to SPFE keep changing, although networks always stabilize at the end. This input diversity to SPFE adds regularization to RSN. Layer normalization \cite{ba2016layer} instead of batch normalization \cite{ioffe2015batch} is used in the PointNet within each voxel because the number of foreground points varies a lot among input frames.

We rely on two widely adopted data augmentation strategies, including random flipping along the X axis and global rotation around the Z axis with a random angle sampled from $[-\pi/4, \pi/4]$ on the selected foreground points.

During inference, we reuse past learned range features and segmentation results (outputs of foreground point selection) such that the inference cost for temporal models remains similar as the single frame models.

\subsection{Results}
All detection results are measured using the official WOD evaluation detection metrics which are BEV and 3D average precision (AP), heading error weighted BEV and 3D average precision (APH) for L1 (easy) and L2 (hard) difficulty levels \cite{sun2020scalability}. The IoU threshold is set as 0.7 for vehicle, 0.5 for pedestrian. We show results on the validation set for all our models in Table \ref{vehicle_result} and Table \ref{ped_result}, results on the official test set in Table \ref{test_set_result}. The latency numbers are obtained on Tesla V100 GPUs with float32 without TensorRT except PVRCNN which is obtained on Titan RTX from PVRCNN authors. In order to better show the latency improvement from our RSN model, NMS timing is not included in all of the baselines because our efficient detection head can be adapted to most of other baselines. We do not show timing for our single frame models as their latency is bounded by their multi-frame correspondences.


Table \ref{vehicle_result} shows that our single frame model CarS\_1f is at least 3x more efficient than the baselines while still being more accurate than all single stage methods. Its temporal version boosts the accuracy further at negligible additional inference costs. CarXL\_3f significantly outperforms all published methods. It also outperforms PVRCNN-WOD \cite{shi2020pvreport}, the most accurate LiDAR only model submitted in the Waymo Open Dataset Challenge.

Table \ref{ped_result} shows more significant improvements on both efficiency and accuracy on pedestrian detection. The efficient single frame model PedS\_1f is significantly more accurate and efficient than all published single-stage baseline models. Its temporal version further improves accuracy. The less efficient model PedL\_3f , outperforms PVRCNN-WOD \cite{shi2020pvreport}, while still being significantly more efficient than all baselines. We see additional efficiency gains on pedestrian detection compared with vehicle detection because there are much fewer people-foreground points. Given the high resolution range image and the high recall foreground segmentation, our model is a great fit for real time small object detection.

Table \ref{test_set_result} shows that RSN ensemble outperforms the PVRCNN WOD challenge submission \cite{shi2020pvreport} which is an ensemble of many models.

\figref{fig:ex} shows a few examples picked from Waymo Open Dataset validation set to demonstrate the model quality in dealing with various hard cases such as a crowd of pedestrians, small objects with few points, large objects, and moving objects in temporal model.

\subsection{Foreground Point Selection Experiments}

\begin{figure}[th!]
    \centering
    \includegraphics[width=0.93\columnwidth]{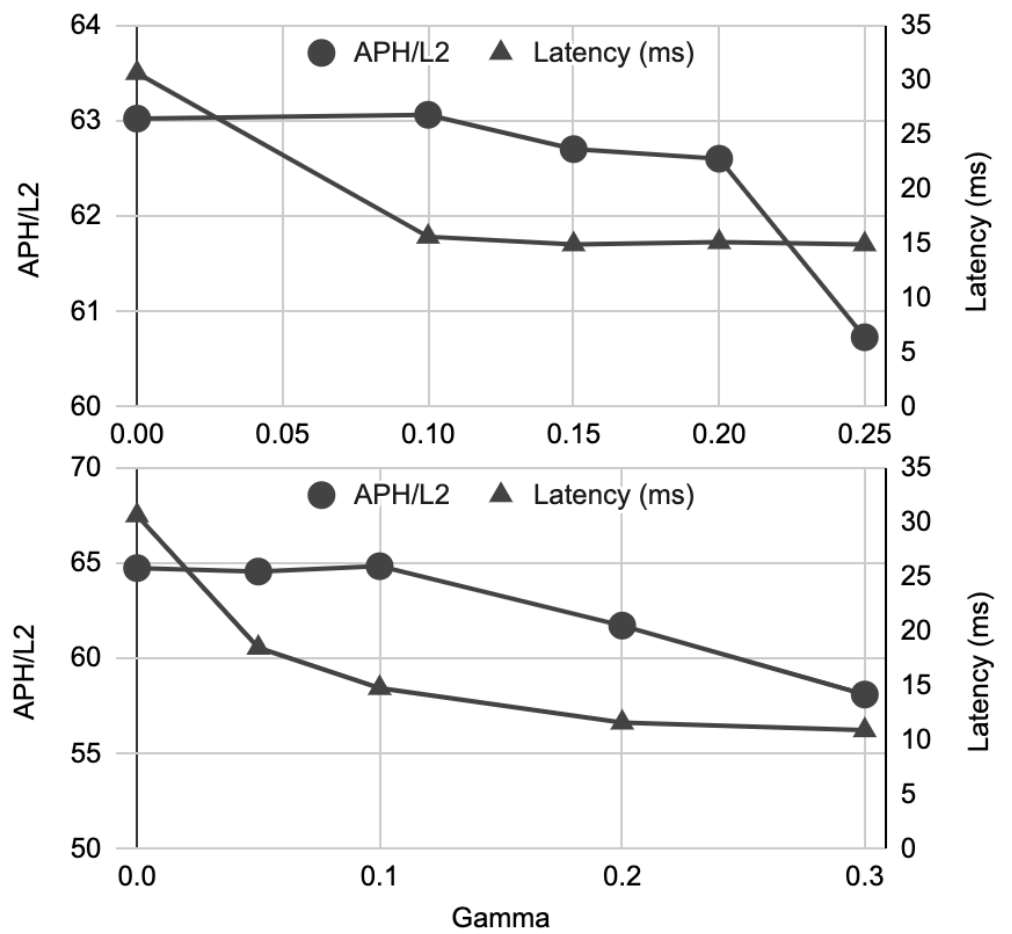}
    \caption{Model performance for different foreground point selection thresholds $\gamma$ as defined in \S\ref{sec:seg}. Top: vehicle result for RSN CarS\_3f. Bottom: pedestrian result for model PedS\_3f. The model accuracy (by APH/L2) does not decrease much but latency drops rapidly when $\gamma$ is less than a certain threshold. \vspace{-0.3cm}}
    \label{fig:ps_study_gamma}
\end{figure}

Foreground point selection is one of the major contributions in the RSN model that supports better efficiency. We conduct experiments by scanning the foreground selection threshold $\gamma$ described in \S\ref{sec:spfe}. As shown in Fig. \ref{fig:ps_study_gamma}, there exists a score threshold $\gamma$ that reduces model latency significantly with negligible impact on accuracy.

In practice, $\gamma$ and $\lambda_1$ in Eq \ref{eq:total_loss} need to be set to values so that selected foreground points have high recall and enough accuracy to achieve good speedup. In our experiments, foreground precision/recall is 77.5\%/99.6\% for CarS\_3f and 15.3\%/97.6\% for PedS\_3f.  We can start with low $\gamma$ and scanning some possible values of $\lambda_1$ to pick one $\lambda_1$. Then we grid search a few $\gamma$.

\subsection{Ablation study}
\label{sec:ab}
In this section, we show additional ablation studies in order to gain insight into model design choices. All experiments in this section are conducted on our efficient models CarS\_3f and on PedS\_3f. 

\begin{table}[th!]
\footnotesize
\begin{center}
\begin{tabular}{l|cc|cc}
\toprule 
        & \multicolumn{2}{c}{Vehicle} & \multicolumn{2}{|c}{Pedestrian} \\
        & AP/APH L2 & Latency & AP/APH L2 & Latency  \\ \midrule
Baseline & 64.2/63.9 & 15.4 & 68.88/66.07 & 14.4 \\ 
-RI & 60.9/60.3 & 27.0 & 63.5/60.5 & 30.0 \\
-E2E & 60.1/59.7 & 15.6 & 64.8/61.7 & 14.6 \\
+xyz  & 64.1/63.7 & - & 64.2/61.3 & - \\
-Norm & 60.6/60.2  & - & 64.7/61.9 & - \\
\bottomrule
\end{tabular}
\end{center}
\caption{The 3D AP/APH at LEVEL\_2 difficulty and latency in milliseconds on the \textit{validation set} for several ablations (\S\ref{sec:ab}). 
\vspace{-0.3cm}}
\label{table:ablation}
\end{table}

Table \ref{table:ablation} shows that features learnt from range image not only help segment foreground points, thus supporting model efficiency, but also improve model accuracy as shown in row \textit{-RI}. Accuracy improvement is higher for pedestrians because of the high resolution semantic feature learned especially impacting the long range. Gradients passed from SPFE to RIFE help detection accuracy as shown in row \textit{-E2E}. Temporal variant features $(x, y, z)$ with ego-motion removed hurt detection accuracy for pedestrian detection as shown in row \textit{+xyz}. Detection accuracy drops if the heatmap normalization described in \S\ref{sec:br} is disabled as shown in row \textit{-Norm}.

\subsection{Scalability}
To further demonstrate RSN's model scalabilty, we conducted experiments on an internal dataset collected from higher quality longer range LiDARs. Here, the detection range is a square of size $[-250m, 250m] \times [-250m, 250m]$ and that is centered at the SDC. This is beyond the memory capacity of PointPillars \cite{lang2019pointpillars} running on a Tesla v100 GPU. We have trained RSN CarS\_3f and a variant with RIFE and foreground point selection removed on this dataset. As shown in table~\ref{table:scalability}, RSN can scale to a significantly larger detection range with good accuracy and efficiency. This demonstrates that foreground sampling and range image features remain effective in the larger detection range.

\begin{table}[th!]
\footnotesize
\begin{center}
\begin{tabular}{c|ccc}
\toprule 
 RIFE stage & BEV APH & 3D APH & Latency (ms)  \\ \midrule
\cmark & 83.6 & 61.2 & 22 \\ 
\xmark & 79.4 & 53.4 & 44 \\
\bottomrule
\end{tabular}
\end{center}
\caption{The APH  and latency in milliseconds on the \textit{test set} of an internal long range LiDAR dataset for vehicle detection with model CarS\_3f. \vspace{-0.5cm}}
\label{table:scalability}
\end{table}

\section{Conclusions}
We have introduced RSN, a novel range image based 3D object detection method that can be trained end-to-end using LiDAR data. The network operates in the large detection range required for safe, fast-speed driving. In the Waymo Open Dataset, we show that RSN outperforms all existing LiDAR-only methods by offering higher detection performance (AP/APH on both BEV and 3D) as well as faster running times. For future work, we plan to explore alternative detection heads and optimized SPFE in order to better take advantage of the sparsity of the foreground points.

{\small
\bibliographystyle{ieee_fullname}
\bibliography{egbib}

\begin{thebibliography}{10}\itemsep=-1pt

\bibitem{tensorflow}
Martin Abadi, Paul Barham, Jianmin Chen, Zhifeng Chen, Andy Davis, Jeffrey
  Dean, Matthieu Devin, Sanjay Ghemawat, Geoffrey Irving, Michael Isard,
  Manjunath Kudlur, Josh Levenberg, Rajat Monga, Sherry Moore, Derek~G. Murray,
  Benoit Steiner, Paul Tucker, Vijay Vasudevan, Pete Warden, Martin Wicke, Yuan
  Yu, and Xiaoqiang Zheng.
\newblock Tensorflow: A system for large-scale machine learning.
\newblock In {\em 12th USENIX Symposium on Operating Systems Design and
  Implementation (OSDI 16)}, pages 265--283, 2016.

\bibitem{ba2016layer}
Jimmy~Lei Ba, Jamie~Ryan Kiros, and Geoffrey~E Hinton.
\newblock Layer normalization.
\newblock {\em arXiv preprint arXiv:1607.06450}, 2016.

\bibitem{bewley2020range}
Alex Bewley, Pei Sun, Thomas Mensink, Dragomir Anguelov, and Cristian
  Sminchisescu.
\newblock Range conditioned dilated convolutions for scale invariant 3d object
  detection.
\newblock In {\em Conference on Robot Learning}, 2020.

\bibitem{chen2017mv3d}
Xiaozhi Chen, Huimin Ma, Ji Wan, Bo Li, and Tian Xia.
\newblock Multi-view 3d object detection network for autonomous driving.
\newblock In {\em Proceedings of the IEEE Conference on Computer Vision and
  Pattern Recognition}, pages 1907--1915, 2017.

\bibitem{cheng2020improving}
Shuyang Cheng, Zhaoqi Leng, Ekin~Dogus Cubuk, Barret Zoph, Chunyan Bai, Jiquan
  Ngiam, Yang Song, Benjamin Caine, Vijay Vasudevan, Congcong Li, et~al.
\newblock Improving 3d object detection through progressive population based
  augmentation.
\newblock In {\em ECCV}, 2020.

\bibitem{dai2016rfcn}
Jifeng Dai, Yi Li, Kaiming He, and Jian Sun.
\newblock R-fcn: Object detection via region-based fully convolutional
  networks.
\newblock In {\em Advances in neural information processing systems}, pages
  379--387, 2016.

\bibitem{engelcke2017vote3deep}
Martin Engelcke, Dushyant Rao, Dominic~Zeng Wang, Chi~Hay Tong, and Ingmar
  Posner.
\newblock Vote3deep: Fast object detection in 3d point clouds using efficient
  convolutional neural networks.
\newblock In {\em 2017 IEEE International Conference on Robotics and Automation
  (ICRA)}, pages 1355--1361. IEEE, 2017.

\bibitem{ge2020afdet}
Runzhou Ge, Zhuangzhuang Ding, Yihan Hu, Yu Wang, Sijia Chen, Li Huang, and
  Yuan Li.
\newblock Afdet: Anchor free one stage 3d object detection.
\newblock {\em arXiv preprint arXiv:2006.12671}, 2020.

\bibitem{girshick2015fast}
Ross Girshick.
\newblock Fast r-cnn.
\newblock In {\em Proceedings of the IEEE international conference on computer
  vision}, pages 1440--1448, 2015.

\bibitem{girshick2014rcnn}
Ross Girshick, Jeff Donahue, Trevor Darrell, and Jitendra Malik.
\newblock Rich feature hierarchies for accurate object detection and semantic
  segmentation.
\newblock In {\em Proceedings of the IEEE conference on computer vision and
  pattern recognition}, pages 580--587, 2014.

\bibitem{graham2017submanifold}
Benjamin Graham and Laurens van~der Maaten.
\newblock Submanifold sparse convolutional networks.
\newblock {\em arXiv preprint arXiv:1706.01307}, 2017.

\bibitem{packnet}
Vitor Guizilini, Rares Ambrus, Sudeep Pillai, Allan Raventos, and Adrien
  Gaidon.
\newblock 3d packing for self-supervised monocular depth estimation.
\newblock In {\em IEEE Conference on Computer Vision and Pattern Recognition
  (CVPR)}, 2020.

\bibitem{HeResnet}
Kaiming He, Xiangyu Zhang, Shaoqing Ren, and Jian Sun.
\newblock Deep residual learning for image recognition.
\newblock In {\em Proceedings of the IEEE Conference on Computer Vision and
  Pattern Recognition (CVPR)}, June 2016.

\bibitem{ioffe2015batch}
Sergey Ioffe and Christian Szegedy.
\newblock Batch normalization: Accelerating deep network training by reducing
  internal covariate shift.
\newblock {\em arXiv preprint arXiv:1502.03167}, 2015.

\bibitem{kingma2014adam}
Diederik~P Kingma and Jimmy Ba.
\newblock Adam: A method for stochastic optimization.
\newblock {\em arXiv preprint arXiv:1412.6980}, 2014.

\bibitem{lang2019pointpillars}
Alex~H Lang, Sourabh Vora, Holger Caesar, Lubing Zhou, Jiong Yang, and Oscar
  Beijbom.
\newblock Pointpillars: Fast encoders for object detection from point clouds.
\newblock In {\em CVPR}, 2019.

\bibitem{law2018cornernet}
Hei Law and Jia Deng.
\newblock Cornernet: Detecting objects as paired keypoints.
\newblock In {\em Proceedings of the European Conference on Computer Vision
  (ECCV)}, pages 734--750, 2018.

\bibitem{liang2021rangercnn}
Zhidong Liang, Ming Zhang, Zehan Zhang, Xian Zhao, and Shiliang Pu.
\newblock Rangercnn: Towards fast and accurate 3d object detection with range
  image representation, 2021.

\bibitem{lin2017focal}
Tsung-Yi Lin, Priya Goyal, Ross Girshick, Kaiming He, and Piotr Doll{\'a}r.
\newblock Focal loss for dense object detection.
\newblock In {\em Proceedings of the IEEE international conference on computer
  vision}, pages 2980--2988, 2017.

\bibitem{liu2016ssd}
Wei Liu, Dragomir Anguelov, Dumitru Erhan, Christian Szegedy, Scott Reed,
  Cheng-Yang Fu, and Alexander~C Berg.
\newblock Ssd: Single shot multibox detector.
\newblock In {\em ECCV}, 2016.

\bibitem{meyer2019lasernet}
Gregory~P Meyer, Ankit Laddha, Eric Kee, Carlos Vallespi-Gonzalez, and Carl~K
  Wellington.
\newblock Lasernet: An efficient probabilistic 3d object detector for
  autonomous driving.
\newblock In {\em CVPR}, 2019.

\bibitem{ngiam2019starnet}
Jiquan Ngiam, Benjamin Caine, Wei Han, Brandon Yang, Yuning Chai, Pei Sun, Yin
  Zhou, Xi Yi, Ouais Alsharif, Patrick Nguyen, et~al.
\newblock Starnet: Targeted computation for object detection in point clouds.
\newblock {\em arXiv preprint arXiv:1908.11069}, 2019.

\bibitem{qi2019deepvote}
Charles~R Qi, Or Litany, Kaiming He, and Leonidas~J Guibas.
\newblock Deep hough voting for 3d object detection in point clouds.
\newblock In {\em Proceedings of the IEEE International Conference on Computer
  Vision}, pages 9277--9286, 2019.

\bibitem{qi2018frustum}
Charles~R Qi, Wei Liu, Chenxia Wu, Hao Su, and Leonidas~J Guibas.
\newblock Frustum pointnets for 3d object detection from rgb-d data.
\newblock In {\em Proceedings of the IEEE conference on computer vision and
  pattern recognition}, pages 918--927, 2018.

\bibitem{qi2017pointnet}
Charles~R Qi, Hao Su, Kaichun Mo, and Leonidas~J Guibas.
\newblock Pointnet: Deep learning on point sets for 3d classification and
  segmentation.
\newblock In {\em CVPR}, 2017.

\bibitem{qi2017pointnet++}
Charles~Ruizhongtai Qi, Li Yi, Hao Su, and Leonidas~J Guibas.
\newblock Pointnet++: Deep hierarchical feature learning on point sets in a
  metric space.
\newblock In {\em NeurIPS}, 2017.

\bibitem{redmon2016you}
Joseph Redmon, Santosh Divvala, Ross Girshick, and Ali Farhadi.
\newblock You only look once: Unified, real-time object detection.
\newblock In {\em Proceedings of the IEEE conference on computer vision and
  pattern recognition}, pages 779--788, 2016.

\bibitem{ren2016faster}
Shaoqing Ren, Kaiming He, Ross Girshick, and Jian Sun.
\newblock Faster r-cnn: Towards real-time object detection with region proposal
  networks.
\newblock {\em IEEE transactions on pattern analysis and machine intelligence},
  39(6):1137--1149, 2016.

\bibitem{ronneberger2015u}
Olaf Ronneberger, Philipp Fischer, and Thomas Brox.
\newblock U-net: Convolutional networks for biomedical image segmentation.
\newblock In {\em MICCAI}, 2015.

\bibitem{shi2020pv}
Shaoshuai Shi, Chaoxu Guo, Li Jiang, Zhe Wang, Jianping Shi, Xiaogang Wang, and
  Hongsheng Li.
\newblock Pv-rcnn: Point-voxel feature set abstraction for 3d object detection.
\newblock In {\em CVPR}, 2020.

\bibitem{shi2020pvreport}
Shaoshuai Shi, Chaoxu Guo, Jihan Yang, and Hongsheng Li.
\newblock Pv-rcnn: The top-performing lidar-only solutions for 3d detection/3d
  tracking/domain adaptation of waymo open dataset challenges.
\newblock {\em arXiv preprint arXiv:2008.12599}, 2020.

\bibitem{shi2019pointrcnn}
Shaoshuai Shi, Xiaogang Wang, and Hongsheng Li.
\newblock Pointrcnn: 3d object proposal generation and detection from point
  cloud.
\newblock In {\em CVPR}, 2019.

\bibitem{simon2018complex}
Martin Simon, Stefan Milz, Karl Amende, and Horst-Michael Gross.
\newblock Complex-yolo: Real-time 3d object detection on point clouds.
\newblock {\em arXiv preprint arXiv:1803.06199}, 2018.

\bibitem{sun2020scalability}
Pei Sun, Henrik Kretzschmar, Xerxes Dotiwalla, Aurelien Chouard, Vijaysai
  Patnaik, Paul Tsui, James Guo, Yin Zhou, Yuning Chai, Benjamin Caine, et~al.
\newblock Scalability in perception for autonomous driving: Waymo open dataset.
\newblock In {\em CVPR}, 2020.

\bibitem{wang2020pillar}
Yue Wang, Alireza Fathi, Abhijit Kundu, David Ross, Caroline Pantofaru, Tom
  Funkhouser, and Justin Solomon.
\newblock Pillar-based object detection for autonomous driving.
\newblock In {\em ECCV}, 2020.

\bibitem{WaymoV5}
Waymo.
\newblock Waymo's 5th generation driver.
\newblock
  \url{https://blog.waymo.com/2020/03/introducing-5th-generation-waymo-driver.html}.

\bibitem{yan2018second}
Yan Yan, Yuxing Mao, and Bo Li.
\newblock Second: Sparsely embedded convolutional detection.
\newblock {\em Sensors}, 2018.

\bibitem{yang2018pixor}
Bin Yang, Wenjie Luo, and Raquel Urtasun.
\newblock Pixor: Real-time 3d object detection from point clouds.
\newblock In {\em Proceedings of the IEEE conference on Computer Vision and
  Pattern Recognition}, pages 7652--7660, 2018.

\bibitem{yang2019std}
Zetong Yang, Yanan Sun, Shu Liu, Xiaoyong Shen, and Jiaya Jia.
\newblock Std: Sparse-to-dense 3d object detector for point cloud.
\newblock In {\em Proceedings of the IEEE International Conference on Computer
  Vision}, pages 1951--1960, 2019.

\bibitem{zhou2019iou}
Dingfu Zhou, Jin Fang, Xibin Song, Chenye Guan, Junbo Yin, Yuchao Dai, and
  Ruigang Yang.
\newblock Iou loss for 2d/3d object detection, 2019.

\bibitem{zhou2019objects}
Xingyi Zhou, Dequan Wang, and Philipp Kr{\"a}henb{\"u}hl.
\newblock Objects as points.
\newblock {\em arXiv preprint arXiv:1904.07850}, 2019.

\bibitem{zhou2019end}
Yin Zhou, Pei Sun, Yu Zhang, Dragomir Anguelov, Jiyang Gao, Tom Ouyang, James
  Guo, Jiquan Ngiam, and Vijay Vasudevan.
\newblock End-to-end multi-view fusion for 3d object detection in lidar point
  clouds.
\newblock In {\em CORL}, 2019.

\bibitem{zhou2018voxelnet}
Yin Zhou and Oncel Tuzel.
\newblock Voxelnet: End-to-end learning for point cloud based 3d object
  detection.
\newblock In {\em CVPR}, 2018.

\end{thebibliography}
}

\clearpage
\newpage

\appendix


\section{Additional details on SPFE}
\label{app:spfe_detail}
\begin{figure}[th!]
    \centering
    \includegraphics[width=1\columnwidth]{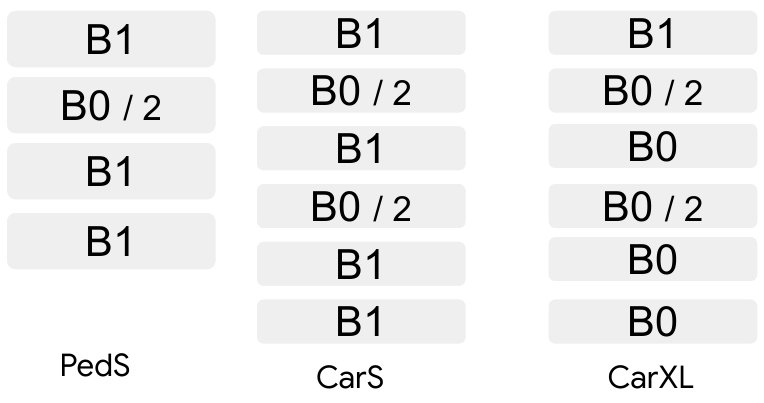}
    \caption{SPFE net architectures for CarS, PedS and CarXL. \vspace{-0.5cm}}
    \label{fig:spfe_more}
\end{figure}

SPFE is composed of blocks illustrated in Fig. \ref{fig:sconv}. PedL and CarL have been illustrated in Fig. \ref{fig:sconv}. Architecture details of PedS, CarS and CarXL can be found in Fig. \ref{fig:spfe_more}. PedS, PedL, CarS, CarL use 2D sparse convolutions and have channel size for all convolutions set to 96. CarXL  use 3D sparse convolutions and have channel size for all convolutions set to 64. CarXL does not have PointNet within each 3D voxel.


\section{More Details on Temporal Fusion}
\begin{figure}[h!]
\begin{center}
   \includegraphics[width=0.85\linewidth]{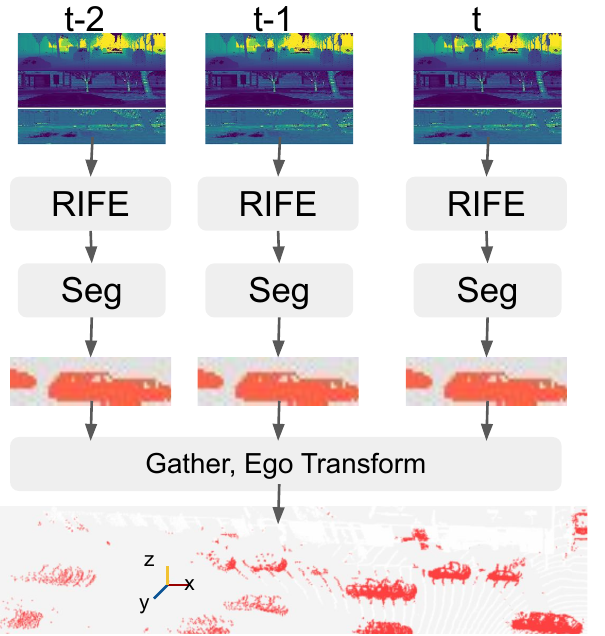}
\end{center}
   \caption{Expanded temporal RSN architecture before SPFE.}
\label{fig:arch_temp}
\end{figure}

1) Temporal RSN duplicates the RIFE (\S \ref{sec:ri_fe}) and Foreground Point Selection part (\S \ref{sec:spfe}) for each temporal frame. Shown in  Fig. \ref{fig:arch_temp}, each branch shares weights and matches the architecture for single frame RSN. These branches are trained together
while during inference only the last frame is computed as other time-steps reuse previous results.
2) After segmentation branches, points are gathered to multiple set of points $P_{\delta_i}$ where $\delta_i$ is the frame time difference between frame 0 (latest frame) and frame $i$ which is usually close to $0.1 * i$ seconds. 
Each point $\boldsymbol{p}$ in $P_{\delta_i}$ is augmented with $\boldsymbol{p - m}, \textbf{\textrm{var}}, \boldsymbol{p - c}$, $\delta_i$, and features learned from RIFE stage where $\boldsymbol{m}$, $\textbf{\textrm{var}}$ is the voxel statistics from $P_{\delta_i}$. After this per frame voxel feature augmentation, all the points are merged to one set $P$ followed by normal voxelization and point net. The rest of the model is the same as single frame models. 3) Given an input sequence $F=\{f_i|i=0, 1,...,\}$, frames are re-grouped into $\Tilde{F} = \{(f_i, f_{i-1}, ..., f_{i-k})|i=0, 1, ...\}$ to train a $k+1$-frame temporal RSN model with target output for frame $i$. If $i-k < 0$, we reuse the last valid frame.

\section{Ensemble Details}
\label{app:ensem}

We provide additional description of the ensembling approach used to produce results highlighted in Table \ref{test_set_result}. We combine both data-level and test-time augmentation-based voting schemes:
We trained five copies of the proposed model, each using a disjoint subset of 80\% of the original training data. For each of the trained model, we perform box prediction under five random point cloud augmentations including random rotation and translation. This procedure yields 25 sets of results in total for each sample. We then use the box aggregation strategy proposed by Solovyev et al.\footnote{Weighted Boxes Fusion: ensembling boxes for object detection models. Solovyev et al.}, extended to 3D boxes with a yaw heading.



\end{document}